\documentclass[runningheads]{llncs}
\usepackage{graphicx}
\usepackage{cite}
\usepackage{xcolor}
\usepackage{tikz}
\usepackage{amsmath,amssymb,amsfonts}
\usepackage{array}
\usepackage[binary-units=true]{siunitx}
\usepackage{pdfpages}
\usepackage{fancyhdr}

\definecolor{tud0d}{cmyk/RGB/HTML}{0,0,0,.8/83,83,83/535353}
\definecolor{tud0c}{cmyk/RGB/HTML}{0,0,0,.6/137,137,137/898989}
\definecolor{tud0b}{cmyk/RGB/HTML}{0,0,0,.4/181,181,181/B5B5B5}
\definecolor{tud0a}{cmyk/RGB/HTML}{0,0,0,.2/220,220,220/DCDCDC}
\definecolor{tud1a}{cmyk/RGB/HTML}{.7,.4,0,0/93,133,195/5D85C3}
\definecolor{tud2a}{cmyk/RGB/HTML}{0.8,.2,0,0/0,156,218/009CDA}
\definecolor{tud3a}{cmyk/RGB/HTML}{0.7,0,.5,0/80,182,149/50B695}
\definecolor{tud4a}{cmyk/RGB/HTML}{.4,0,.8,0/175,204,80/AFCC50}
\definecolor{tud5a}{cmyk/RGB/HTML}{.2,0,.8,0/221,223,72/DDDF48}
\definecolor{tud6a}{cmyk/RGB/HTML}{0,.1,.7,0/255,224,92/FFE05C}
\definecolor{tud7a}{cmyk/RGB/HTML}{0,.3,.8,0/248,186,60/F8BA3C}
\definecolor{tud8a}{cmyk/RGB/HTML}{0,.6,.8,0 /238,122,52/EE7A34}
\definecolor{tud9a}{cmyk/RGB/HTML}{0,.8,.7,0/233,80,62/E9503E}
\definecolor{tud10a}{cmyk/RGB/HTML}{.2,.9,0,0/201,48,142/C9308E}
\definecolor{tud11a}{cmyk/RGB/HTML}{.6,.8,0,0/128,69,151/804597}
\definecolor{tud1b}{cmyk/RGB/HTML}{1,.6,0,0/0,90,169/005AA9}
\definecolor{tud2b}{cmyk/RGB/HTML}{1,.3,0,0/0,131,204/0083CC}
\definecolor{tud3b}{cmyk/RGB/HTML}{1,0,.6,0/0,157,129/009D81}
\definecolor{tud4b}{cmyk/RGB/HTML}{.5,0,1,0/153,192,0/99C000}
\definecolor{tud5b}{cmyk/RGB/HTML}{.3,0,1,0/201,212,0/C9D400}
\definecolor{tud6b}{cmyk/RGB/HTML}{0,.2,1,0/253,202,0/FDCA00}
\definecolor{tud7b}{cmyk/RGB/HTML}{0,.4,1,0/245,163,0/F5A300}
\definecolor{tud8b}{cmyk/RGB/HTML}{0,.7,1,0/236,101,0/EC6500}
\definecolor{tud9b}{cmyk/RGB/HTML}{0,1,.9,0/230,0,26/E6001A}
\definecolor{tud10b}{cmyk/RGB/HTML}{.4,1,0,0/166,0,132/A60084}
\definecolor{tud11b}{cmyk/RGB/HTML}{.7,1,0,0/114,16,133/721085}
\definecolor{tud1c}{cmyk/RGB/HTML}{1,.7,.2,0/0,78,138/004E8A}
\definecolor{tud2c}{cmyk/RGB/HTML}{1,.5,.2,0/0,104,157/00689D}
\definecolor{tud3c}{cmyk/RGB/HTML}{1,.2,.6,0/0,136,119/008877}
\definecolor{tud4c}{cmyk/RGB/HTML}{.6,.1,1,0/127,171,22/7FAB16}
\definecolor{tud5c}{cmyk/RGB/HTML}{.4,.1,1,0/177,189,0/B1BD00}
\definecolor{tud6c}{cmyk/RGB/HTML}{.2,.3,1,0/215,172,0/D7AC00}
\definecolor{tud7c}{cmyk/RGB/HTML}{.2,.5,1,0/210,135,0/D28700}
\definecolor{tud8c}{cmyk/RGB/HTML}{.2,.8,1,0/204,76,3/CC4C03}
\definecolor{tud9c}{cmyk/RGB/HTML}{.3,1,.9,0/185,15,34/B90F22}
\definecolor{tud10c}{cmyk/RGB/HTML}{.5,1,.3,0/149,17,105/951169}
\definecolor{tud11c}{cmyk/RGB/HTML}{.8,1,.2,0/97,28,115/611C73}
\definecolor{tud1d}{cmyk/RGB/HTML}{1,.9,.3,0/36,53,114/243572}
\definecolor{tud2d}{cmyk/RGB/HTML}{1,.7,.4,0/0,78,115/004E73}
\definecolor{tud3d}{cmyk/RGB/HTML}{1,.4,.7,0/0,113,94/00715E}
\definecolor{tud4d}{cmyk/RGB/HTML}{.7,.3,1,0/106,139,55/6A8B22}
\definecolor{tud5d}{cmyk/RGB/HTML}{.5,.2,1,0/153,166,4/99A604}
\definecolor{tud6d}{cmyk/RGB/HTML}{.4,.4,1,0/174,142,0/AE8E00}
\definecolor{tud7d}{cmyk/RGB/HTML}{.3,.6,1,0/190,111,0/BE6F00}
\definecolor{tud8d}{cmyk/RGB/HTML}{.4,.8,1,0/169,73,19/A94913}
\definecolor{tud9d}{cmyk/RGB/HTML}{.5,1,.9,0/156,28,38/961C26}
\definecolor{tud10d}{cmyk/RGB/HTML}{.7,1,.5,0/115,32,84/732054}
\definecolor{tud11d}{cmyk/RGB/HTML}{.9,1,.3,0/76,34,106/4C226A}

\usepackage[pdfauthor={Reich and Prangemeier}, hidelinks, breaklinks=true]{hyperref}

\newcommand \colorindicator[2]{%
	#1 {\textcolor{#2}{$\blacksquare\!\!\blacksquare$}}%
}

\begin{document}
\title{Multi-StyleGAN: Towards Image-Based Simulation of Time-Lapse Live-Cell
	Microscopy} 
	\titlerunning{Multi-StyleGAN}
	
	\author{Christoph Reich\thanks{Christoph Reich and Tim Prangemeier --- both authors contributed equally}	\and
		Tim Prangemeier\textsuperscript{$\star$} \and
		Christian Wildner \and
		Heinz Koeppl
	}
	
	\authorrunning{C. Reich et al.}

	\institute{Centre for Synthetic Biology,\\Department of Electrical Engineering and Information Technology,\\Department of Biology,\\Technische Universit\"at Darmstadt\\ 
		\email{heinz.koeppl@bcs.tu-darmstadt.de}
	}	
	\maketitle              	

	\thispagestyle{plain}
	\fancypagestyle{plain}{
		\fancyhf{} 
		\fancyfoot[L]{\scriptsize \copyright 2021 Christoph Reich, Tim Prangemeier, Christian Wildner and Heinz Koeppl\\ Revised version accepted to MICCAI 2021 \\ The final authenticated version of this manuscript is published in Lecture Notes in Computer Science, Medical Image Computing and Computer Assisted Intervention - MICCAI 2021 (DOI \url{https://doi.org/10.1007/978-3-030-87237-3_46}).}
		\renewcommand{\headrulewidth}{0pt}
		\renewcommand{\footrulewidth}{0pt}
	}

	\begin{abstract}
		Time-lapse fluorescent microscopy (TLFM) combined with predictive mathematical modelling is a powerful tool to study the inherently dynamic processes of life on the single-cell level. Such experiments are costly, complex and labour intensive. A complimentary approach and a step towards \textit{in silico} experimentation, is to synthesise the  imagery itself. Here, we propose Multi-StyleGAN as a descriptive approach to simulate time-lapse fluorescence microscopy imagery of living cells, based on a past experiment. This novel generative adversarial network synthesises a multi-domain sequence of consecutive timesteps. We showcase Multi-StyleGAN on imagery of multiple live yeast cells in microstructured environments and train on a dataset recorded in our laboratory. The simulation captures underlying biophysical factors and time dependencies, such as cell morphology, growth, physical interactions, as well as the intensity of a fluorescent reporter protein. An immediate application  is to generate additional training and validation data for feature extraction algorithms or to aid and expedite development of advanced experimental techniques such as online monitoring or control of cells. 

		Code and dataset is available at \url{https://git.rwth-aachen.de/bcs/projects/tp/multi-stylegan}.
		
	\end{abstract}
	
	\section{Introduction} \label{sec:inctroduction}
\begin{figure}[h!]
		\centering
		\includegraphics{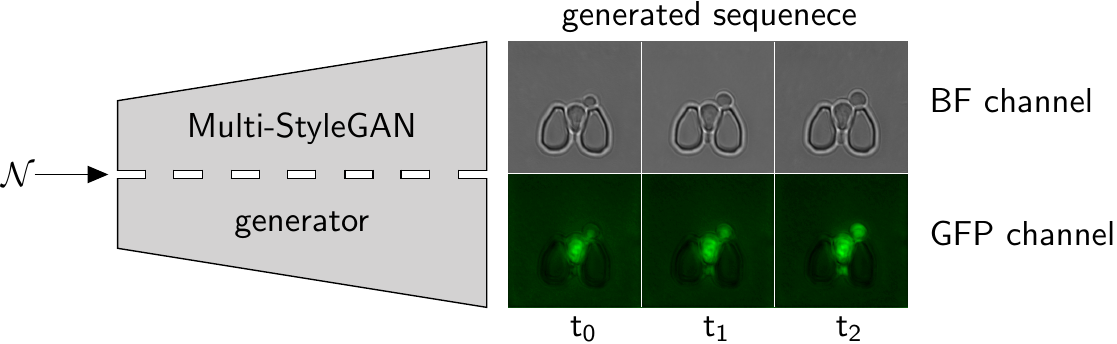}
		\caption{Multi-StyleGAN simulation of yeast TLFM imagery consisting of three consecutive multi-domain timesteps; brightfield top row, fluorescent channel bottom row.}
		\label{fig:figure_1}
	\end{figure}

	Time-lapse fluorescent microscope (TLFM) is a powerful tool to study the inherently dynamic processes of life on the single-cell level \cite{Leygeber2019,Lugagne2020,Pepperkok2006,Prangemeier2020,Chessel2019}. TLFM yields vast amounts of multi-domain imagery from which pertinent quantitative measures can be extracted.  These domains are typically a brightfield (BF) channel that captures the spatial structure and organisation of cells (Fig. \ref{fig:figure_1} top), and one or more fluorescent channels (Fig. \ref{fig:figure_1} bottom) upon which the abundance of biomolecular species can be quantified from fluorescence intensities \cite{Prangemeier2020,Lugagne2020,Leygeber2019,Prangemeier2020b,Prangemeier2020c}. These quantitative measures promise to constitute the backbone for understanding and \textit{de novo} design of biomolecular functionality with explanatory and predictive mathematical models in systems and synthetic biology \cite{Pepperkok2006,Wang2020,Henningsen2020,Prangemeier2020}. Ideally, computer-aided engineering of biological systems will become as routine and reliable as it is today for mechanical or electrical systems, for example.
	
	TLFM experiments yield valuable high-throughput time-lapse fluorescence data on the single-cell level, however, they are costly, labour intensive, and complex \cite{Prangemeier2020,Lugagne2020,Leygeber2019}. A complimentary approach to predictive modelling of the pertinent features extracted from these experiments, and a further step towards \textit{in silico} experimentation, is to simulate experiments by synthesising the imagery itself. While this approach is primarily descriptive in nature, it may be able to capture the broader context of cell morphology and the spatio-temporal structure of multiple cells, or other biophysical features which are not routinely extracted from the imagery. In the future,  interfacing quantitatively predictive modelling of biomolecular circuitry and the spatio-temporal description of multi-cell behaviour is expected to advance our ability to engineer more complex biological microsystems and biomaterials \cite{Hall2020,Ulman2016}. More immediate applications for synthetic microscope imagery are as a means to generate additional data for training and validating of feature extraction algorithms, or to aid and expedite development of advanced experimental techniques such as online monitoring or control of cells \cite{Ulman2016,Bandiera2018,Prangemeier2018,Prangemeier2020b,Prangemeier2020c}.
	
	Generative adversarial networks (GANs) are a recent approach to synthesise images \cite{Goodfellow2014}. They implicitly learn a high-dimensional dataset distribution through unsupervised adversarial training where a \textit{generator} and a \textit{discriminator} play a minimax game \cite{Goodfellow2014}. While the generator synthesises images, the discriminator distinguishes between synthetic \textit{fake} images and \textit{real} images from a training set. GANs have been employed to synthesise a wide range of imagery, such as handwriting, paintings, medical imagery, natural images and faces \cite{Goodfellow2014,Karras2020b}. StyleGAN2 is the current state-of-the-art for high-resolution images \cite{Karras2020a}. 
	
	The generation of synthetic cell imagery dates back to the late 1990s \cite{Ulman2016}. Recently GANs have been employed, for example, to synthesise fluorescent microscope images of isolated \textit{Schizosaccharomyces pombe} or human cells in the centre of the frame \cite{Osokin2017,Johnson2017,Goldsborough2017}. Synthetic images of multiple blood cells were generated for data augmentation with conditional GANs \cite{Bailo2019}. GANs have also been employed to infer one microscope modality, such as fluorescence or enhanced contrast imagery from another modality \cite{Han2017,Wieslander2021,Lee2018,Lee2021} or to increase image spatial resolution \cite{Zhang2019}. The spatial organisation of tissue on electron microscopy imagery has been simulated with supervised GANs\cite{Han2018}. The interpolation of video frames between recorded TLFM timesteps has also recently been demonstrated \cite{Comes2020}. To date, we are not aware of any GAN simulations of brightfield imagery of multiple yeast cells, nor of any simulations that capture the growth and spatio-temporal development of cells in future timestep sequences. 
	
	In this study, we propose Multi-StyleGAN to synthesise sequences of multi-domain TLFM imagery of multiple yeast cells in microstructured environments. We introduce a novel dual-styled-convolutional block with separate convolutional paths for each domain. This enables the Multi-StyleGAN generator to learn multi-domain microscope images. We present the corresponding TLFM dataset recorded in our laboratory. Both the brightfield and a fluorescent channel are simulated at three consecutive timesteps. Dynamic behaviour such as changes in morphology, cell growth, their movement, their mechanical interactions with each other and the environment are captured. To the best of our knowledge, this is the first GAN to synthesise brightfield and fluorescence yeast microscopy, the first to simulate multiple yeast cells, as well as the first simulation over multiple timesteps. 
	
	\section{Dataset} \label{sec:dataset}
\begin{figure}[htbp]
		\centerline
		{\includegraphics[width=1\columnwidth]{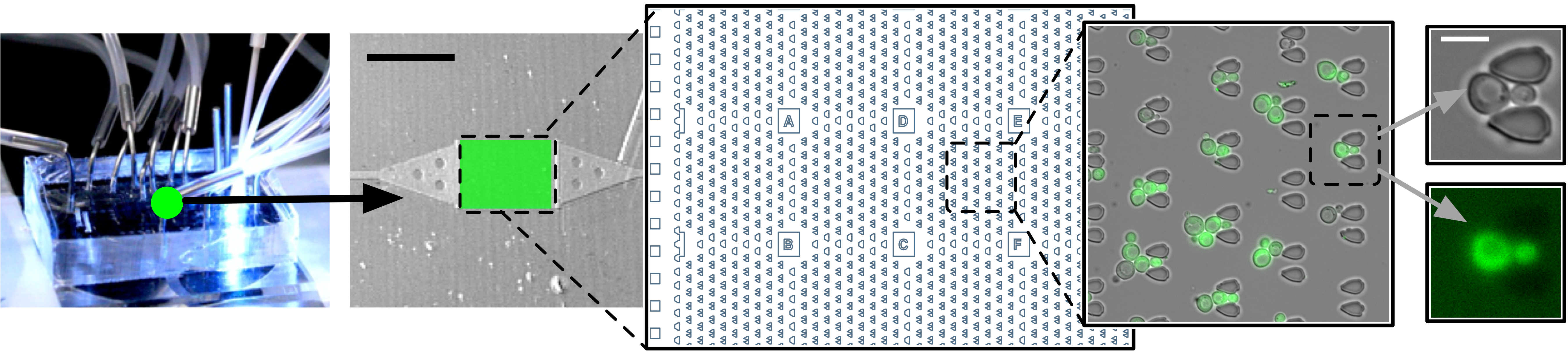}}
		\caption{TLFM setup. Microfluidic chip on microscope table (left). The imaging chamber (green rectangle) contains an array of approximately \SI{1000}{} traps. Overlay of brightfield and fluorescent channel showing fluorescent cells in traps. A brightfield sample with a pair of trap microstructures and two yeast cell, as well as the corresponding fluorescent channel sample (right). Black scale bar \SI{1}{\milli\meter}, white scale bar \SI{10}{\micro\meter}.} 
		\label{fig:complex}
	\end{figure}

	Optical access to living cells is generally enabled by confining these to a monolayer within the focal plane of a microscope (Fig. \ref{fig:complex}). The monolayer is achieved by loading cells into a gap approximately the size of a cell diameter between a cover slip and microstructured polydimethylsiloxane\cite{Prangemeier2020,Leygeber2019}. In the microfluidic configuration we consider here, the microchip is perfused with a constant flow of yeast growth media and maintained at temperatures conducive to yeast growth. The cells are hydrodynamically trapped in the microstructures, constraining these horizontally \cite{Prangemeier2020,Leygeber2019}. The flow enables long term imaging of up to several days, by removing daughter cells and preventing chip crowding. Examples of the routine employ of this configuration include Fig. \ref{fig:complex} and \cite{Prangemeier2020b,Prangemeier2020,Hofmann2019,Leygeber2019,Prangemeier2020c}. 
	
	The training dataset was recorded from one yeast TLFM experiment in our laboratory. The dataset is structured in sequences of at least nine timesteps and includes slight variations in focal plane. Images were selected to each contain less than twelve cells, the majority of which remain inside the frame throughout the sequence. The dataset includes 9696 images of both brightfield and green fluorescent protein (GFP) channels at a resolution of $256\times 256$ (Fig. \ref{fig:complex} (right) and Fig. \ref{fig:gan_results} (left)). 8148 sequences are available to train Multi-StyleGAN when utilising overlapping sequences of three images. 
	
	\section{Methodology} \label{sec:method}
\begin{figure}[!ht]
		\centering
		\includegraphics{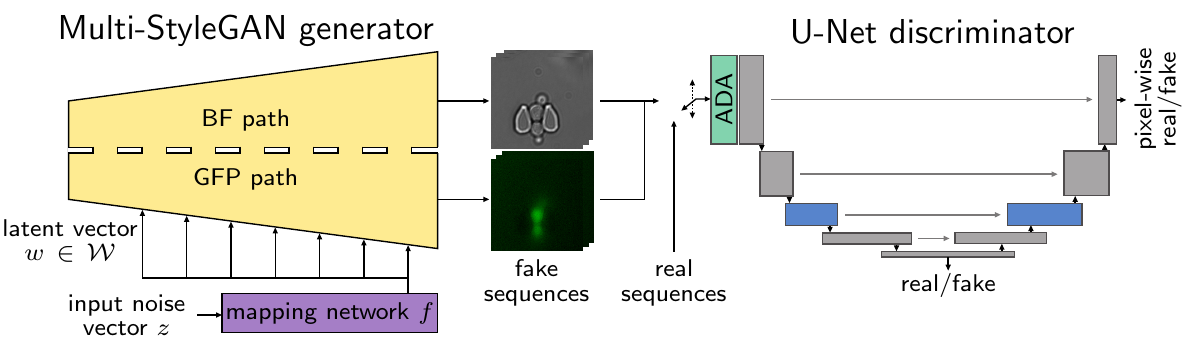}
		\caption{Architecture of Multi-StyleGAN. The style mapping network $f$ (in \colorindicator{purple}{tud11a!60}) transforms the input noise vector $z\sim\mathcal{N}_{512}\left(0, 1\right)$ into a latent vector $w\in\mathcal{W}$, which in turn is passed to each stage of the generator (in \colorindicator{yellow}{tud6a!60}) by three dual-styled-convolutional blocks (Fig. \ref{fig:dual_styled_conv_block}). The generator predicts a sequence of three consecutive images for both the BF and GFP channels. The U-Net discriminator with ADA distinguishes between real and a fake sequences by making both a scalar and a pixel-wise real/fake prediction. Residual discriminator blocks in \colorindicator{gray}{tud0b} and non-local blocks \cite{Wang2018} in \colorindicator{blue}{tud1a}.}
		\label{fig:twin_stylegan}
	\end{figure}

	We propose Multi-StyleGAN (Fig. \ref{fig:twin_stylegan}) for high-resolution ($256^2$) multi-domain image sequence generation. The architecture is influenced by the recent StyleGAN2 \cite{Karras2020b} and star-shaped GAN \cite{Osokin2017}. The latter utilises a generator with two convolutional paths to synthesise a low-resolution ($48\times 80$) two-domain image. We applied this idea to the StyleGAN2 architecture to develope Multi-StyleGAN. 
	
	Initially, we naively adapted StyleGAN2 for sequences of multi-domain imagery, which became the basis of the baselines in this study. Both domains and the time dimension were modeled in the channel dimension. We also employed a StyleGAN2 with 3D convolutions. StyleGAN2 3D models the time dimension in the third convolution dimension. The GFP and BF domains were modeled in the channel dimension. However, even with the use of a U-Net discriminator \cite{Schonfeld2020} and adaptive discriminator augmentation (ADA) \cite{Karras2020b}, these only converged to equilibria with poor generative performance. Samples for the StyleGANs with the best convergence are depicted in Fig. S2 (supplement). We modified the architecture resulting in Multi-StyleGAN, as the StyleGAN2 and StyleGAN2 3D samples are qualitatively unrealistic and not biophysically sensible, in particular for the fluorescent domain which bears a strong resemblance to the BF.
	
	The Multi-StyleGAN generator utilises a mapping network $f$ and two separate 2D convolutional paths, conditioned on the latent vector $w$, to generate a matching BF and GFP image sequence (Fig. \ref{fig:twin_stylegan}). The time dimension is modeled within the feature dimension. A U-Net \cite{Schonfeld2020} serves as the Multi-StyleGAN discriminator network, returning both a scalar and pixel-wise real/fake prediction. This reinforces local and global coherence in the synthesised imagery \cite{Schonfeld2020}. 
	
\begin{figure}[ht]
		\centering
		\includegraphics{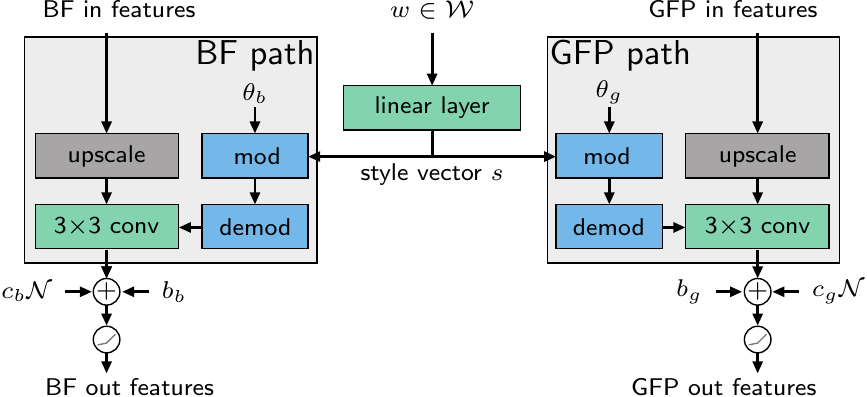}
		\caption{Dual-styled-convolutional block of Multi-StyleGAN. The incoming latent vector $w$ is transformed into the style vector $s$ by a linear layer. This style vector modulates (mod) \cite{Karras2020a} the convolutional weights $\theta_{b}$ and $\theta_{g}$, which are optionally demodulated (demod) \cite{Karras2020a} before convolving the (optionally bilinearly upsampled) incoming features of the previous block. Learnable biasses ($b_b$ and $b_g$) and channel-wise Gaussian noise ($\mathcal{N}$) scaled by a learnable constant ($c_b$ and $c_g$), are added to the features. The final output features are obtained by applying a leaky ReLU activation.}
		\label{fig:dual_styled_conv_block}
	\end{figure}

	The dual-styled-convolutional (DSC) block is the main component of the Multi-StyleGAN generator. It uses two separate convolutional paths (Fig. \ref{fig:dual_styled_conv_block} BF/GFP path) to generate the BF and GFP domains separately. A single style vector modulates \cite{Karras2020a} the convolutional weights of both paths, enforcing consistency between the domains. Multi-StyleGAN utilises three DSC blocks in each of the seven resolution stages. Similarly to the StyleGAN2 output skip architecture \cite{Karras2020a}, two blocks build the main path, and one serves as the output mapping.
	
	Multi-StyleGAN trains unsupervised on the top-$k$ \cite{Sinha2020} non-saturating GAN loss \cite{Goodfellow2014} for both the scalar and pixel-wise prediction of the U-Net discriminator \cite{Schonfeld2020}. Similarly to the original StyleGAN2 training process, path length \cite{Karras2020a} and $R_{1}$ \cite{Mescheder2018} regularization are employed in a lazy fashion \cite{Karras2020a}. Additionally, CutMix augmentation and consistency regularization \cite{Schonfeld2020} is applied to the U-Net discriminator. To enforce learning of time dependencies, real disordered sequences are fed to the discriminator as fake samples. We emloyed ADA \cite{Karras2020b} to prevent the discriminator from overfitting. Due to the used dataset characteristics, only pixel blitting and geometric transformations are applied as augmentations.
	
	We employ the Inception Score \cite{Salimans2016} (IS), Fréchet Inception Distance \cite{Heusel2017} (FID) and Fréchet Video Distance \cite{Unterthiner2019} (FVD) as quantitative metrics to analyse Multi-StyleGANs performance and to facilitate future comparisons. These widespread metrics measure image quality and diversity relative to the training dataset. Technically, the FID measures the similarity between the generated distribution and the dataset distribution in the Inception-Net latent space \cite{Heusel2017}. FVD is the related measure for sequences \cite{Unterthiner2019}. One frame was sampled uniformly from the predicted sequence to compute both the IS and the FID. A trained Inception-Net V3 \cite{Szegedy2016} provided by  Torchvision\footnote{\url{https://github.com//vision}} predicted the statistics for the FID and the IS. We utilised a trained I3D network\footnote{\url{https://github.com/piergiaj/pytorch-i3d}} \cite{Carreira2017} to compute the FVD \cite{Unterthiner2019}. All validation metrics were computed over the whole dataset length (8148 sequences). While these are the most widespread and suitable metrics available, they have some limitations for the scenario studied here \cite{Barratt2018,Karras2020b}. The FID tends to be dominated by an inherent bias for limited real samples \cite{Karras2020b}. Both the Inception-Net and the I3D network are trained on natural images or videos, respectively \cite{Salimans2016, Heusel2017, Unterthiner2019}. These may not fully capture the domain-specific features of the trapped yeast cell dataset, in particular for the fluorescent channel. 
	
	We implemented Multi-StyleGAN using PyTorch \cite{Paszke2019}, and ADA with Kornia \cite{Riba2020}. Each of the seven generator stages  employs $512$ features. The mapping network $f$ is an eight-layered fully connected neural network. The input to $f$ is a $512$-dimensional input noise vector. The U-Net discriminator encoder consists of five blocks with $128$, $256$, $384$, $768$, and $1024$ features. The decoder employs $768$, $384$, $256$, and $128$ features in each respective block. We trained Multi-StyleGAN for $100$ epochs with Adam optimizer \cite{Kingma2015} and the hyperparameters $\beta_{1}=0$, $\beta_{2}=0.99$. The generator and discriminator learning rates were $2\cdot 10^{-4}$ and $6\cdot 10^{-4}$. Exponential-moving-average of the generator weights were used. The learning rate for the mapping network was $2\cdot 10^{-6}$. Training took approximately one day on four Nvidia Tesla V100 ($32\si{\giga\byte}$) with a batch size of $24$. An overview of all hyperparameters is given in the supplement (Table S1).
	
	\section{Results} \label{sec:results}
\begin{figure}[!ht]
		\centering
		\includegraphics{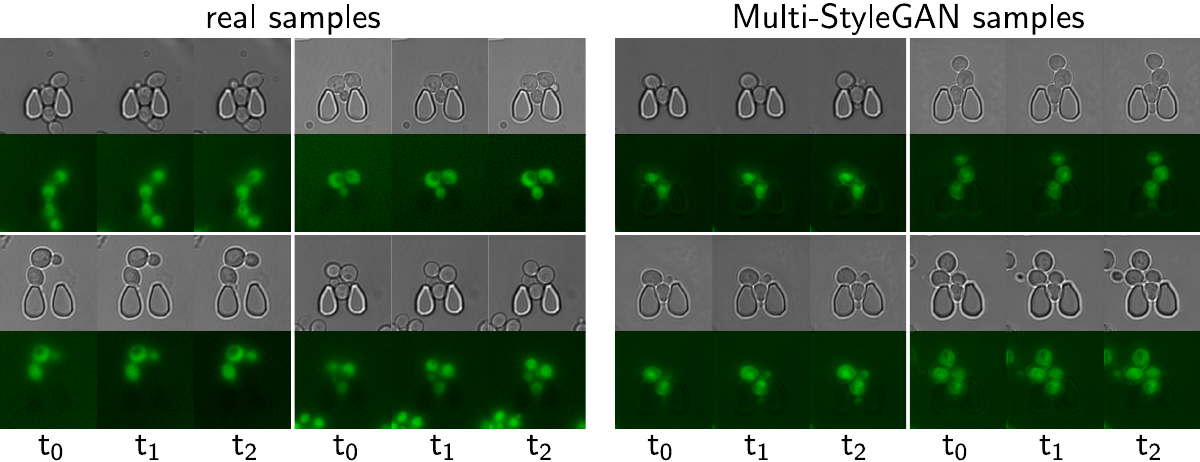}
		\caption{Real BF \& GFP sequences, three timesteps (left). Multi-StyleGAN generated BF \& GFP sequences (right). BF channel in grayscale and GFP channel in green.}
		\label{fig:gan_results}
	\end{figure}

	We demonstrate Multi-StyleGAN's performance at synthesising sequences of consecutive multi-domain TLFM time-points by simulating yeast cells in microstructured environments. Sample sequences are depicted in Fig. \ref{fig:gan_results} (right) and Fig. S2 (supplement). In the brightfield domain, the network successfully captures the microstructures at the correct positions as well as multiple cells at various stages of growth and cell cycle. Cell growth is most evident in the newly budded daughter cells, and as expected biophysically, growth slows for larger cells. Cell fluorescence, and changes thereof, is exhibited on the corresponding channel. Both the BF and GFP domains are aligned. 
	
	In addition to capturing the biophysical features and time dependencies, Multi-StyleGAN synthesises image sequences with a high degree of variation in cell and trap configurations. The fine texture of the cells is captured. The generated samples include slight variations in microscope focus between sequences, leading to light or dark outer \textit{halos} around the cell contours (Fig. \ref{fig:gan_results}). The GAN samples show a similar distribution of these \textit {halos} or focus variations. 
	
	We consider quantitative metrics to analyse Multi-StyleGAN's performance and to facilitate future comparisons (Table \ref{tab:gan_results}). Multi-StyleGAN yields better scores than StyleGAN2 and StyleGAN 3D, in both FID and FVD, on both domains. This is in agreement with visual assessment of the achieved results (supplement Fig. S2). The BF domain achieves significantly better scores for all methods, in all but one case. This may be caused by a mismatch between the domain-specific features of the GFP channel and the networks used to evaluate the metrics that are trained on natural imagery or videos. Multi-StyleGAN achieved an IS of 1.864 and 2.437 for the BF and GFP channel, respectively. Both scores are close to the dataset IS of 2.021 for the BF channel and 2.479 for the GFP channel. This supports the qualitative observation that the imagery generated by Multi-StyleGAN are sharp and diverse in comparison to the dataset.
	
	\begin{table}[!ht]
		\caption{Evaluation metrics for Multi-StyleGAN and baselines.}
		\centering
		\label{tab:gan_results}
		\begin{tabular}{>{\raggedright\arraybackslash}p{0.45\columnwidth}|>{\raggedright\arraybackslash}p{0.125\columnwidth}|>{\raggedright\arraybackslash}p{0.125\columnwidth}|>{\raggedright\arraybackslash}p{0.125\columnwidth}|>{\raggedright\arraybackslash}p{0.125\columnwidth}}
			\hline
			& \multicolumn{2}{l|}{FID\cite{Heusel2017} $\downarrow$} & \multicolumn{2}{l}{FVD\cite{Unterthiner2019} $\downarrow$} \\
			\cline{2-5}
			Method & BF & GFP & BF & GFP \\
			\hline
			Multi-StyleGAN (ours) & $\mathbf{33.3687}$ & $\mathbf{207.8409}$ & $\mathbf{4.4632}$ & $\mathbf{30.1650}$ \\
			StyleGAN2 + ADA + U-Net dis. & $200.5408$ & $224.7860$ & $45.6296$ & $35.2169$ \\
			StyleGAN2 3D + ADA + U-Net dis. & $76.0344$ & $298.7545$ & $14.7509$ & $31.4771$ \\
			\hline
		\end{tabular}
	\end{table}
	
	Multi-StyleGAN utilises two separate convolutional paths in the generator. This decouples the convolutional weights, which are subsequently learnt for the individual domains. The transitions between images are smooth when interpolating through the latent space (see supplementary video), indicating the network does not merely memorise the training data. Multi-StyleGAN preserves the style mixing property of StyleGAN \cite{Karras2019}, allowing images to be manipulated in the latent space \cite{Abdal2020}. This is demonstrated in Fig. S1 (supplement) where the latent vectors of two samples are mixed at different stages of the generator. 
	
	\section{Discussion} \label{sec:discussion}
	The proposed Multi-StyleGAN successfully synthesises a multi-domain microscope imagery sequence of live cells at three consecutive timesteps. Both the brightfield and fluorescent channels capture underlying biophysical factors realistically (Fig. \ref{fig:gan_results}). While some cells bud within the simulated sequence, longer time series over \SI{9}{} timesteps, corresponding to the doubling time of yeast or more are needed to fully capture and simulate the complete cell cycle including the budding process. In the future, this limitation could be counteracted by adapting Multi-StyleGAN for training on longer sequences.	
	
	The trained Multi-StyleGAN we present can be applied to a range of scenarios. A typical application for synthesised microscope imagery is as a data augmentation technique to train feature extraction algorithms \cite{Chessel2019,Ulman2016} such as  cell segmentation tools \cite{Prangemeier2020b,Lugagne2020,Prangemeier2020,Prangemeier2020c}. The simulations of consecutive TLFM timesteps themselves can be employed as \textit{in silico} experiments, for example to develop advanced experimental techniques such as online monitoring of cells or optimal experimental design techniques with cell segmentation in the loop \cite{Ulman2016,Prangemeier2018,Bandiera2018,Chessel2019}. 
	
	Currently, Multi-StyleGAN learns an implicit high-dimensional representation of a single experiment. A promising direction for future research is to extend our approach to a whole campaign of experiments. This would
	allow generating image sequences conditioned on given experimental parameters such as organism type or temperature. In a further step towards \textit{in silico} TLFM experimentation, the descriptive simulations Multi-StyleGAN offers may be interfaced with explanatory and predictive models of specific biomolecular circuitry.
	
	\section{Conclusion} \label{sec:conclusion}
	
	In summary, we propose Mult-StyleGAN to synthesise multi-domain image sequences and showcase it by simulating TLFM imagery \textit{in silico}. To the best of our knowledge, this is the first network to simulate temporal sequences of yeast brightfield imagery, in particular with multiple cells in a microstructured environment. Trained on the presented dataset, the simulations capture the spatio-temporal organisation of multiple yeast cells. Biophysical factors and time-dependencies, such as cell morphology and growth, are realistically simulated concurrently to the cell fluorescence. Immediate applications for Multi-StyleGAN are to generate additional training data for segmentation algorithms or to expedite the development of advanced experimental techniques such as optimal experimental design. While the Multi-StyleGAN simulations are descriptive in nature, they are a step towards more complete \textit{in silico} experimentation, especially if interfaced with predictive mathematical models in the future. 
	
	\subsubsection*{Acknowledgements.} 
	We thank Markus Baier for aid with the computational setup, Klaus-Dieter Voss for aid with the microfluidics fabrication, and Tim Kircher, Tizian Dege, and Florian Schwald for aid with the data preparation.\\
	This work was supported by the Landesoffensive f\"{u}r wissenschaftliche Exzellenz as part of the LOEWE Schwerpunkt CompuGene. H.K. acknowledges support from the European Research Council (ERC) with the consolidator grant CONSYN (nr. 773196).

	\bibliographystyle{splncs04}
	\bibliography{bib}

\begin{thebibliography}{10}
\providecommand{\url}[1]{\texttt{#1}}
\providecommand{\urlprefix}{URL }
\providecommand{\doi}[1]{https://doi.org/#1}

\bibitem{Abdal2020}
Abdal, R., Qin, Y., Wonka, P.: {Image2StyleGAN++}: How to edit the embedded
  images? In: CVPR. pp. 8296--8305 (2020)

\bibitem{Bailo2019}
Bailo, O., Ham, D., Shin, Y.M.: Red blood cell image generation for data
  augmentation using conditional generative adversarial networks. In: CVPRW
  (2019)

\bibitem{Bandiera2018}
Bandiera, L., Hou, Z., Kothamachu, V.B., Balsa-Canto, E., Swain, P.S.,
  Menolascina, F.: On-line optimal input design increases the efficiency and
  accuracy of the modelling of an inducible synthetic promoter. Processes
  \textbf{6}(9) (2018)

\bibitem{Barratt2018}
Barratt, S., Sharma, R.: A note on the inception score. In: ICML Workshop
  (2018)

\bibitem{Carreira2017}
Carreira, J., Zisserman, A.: Quo vadis, action recognition? a new model and the
  kinetics dataset. In: CVPR. pp. 6299--6308 (2017)

\bibitem{Chessel2019}
Chessel, A., {Carazo Salas}, R.E.: {From observing to predicting single-cell
  structure and function with high-throughput/high-content microscopy}. Essays
  Biochem.  \textbf{63}(2),  197--208 (2019)

\bibitem{Comes2020}
Comes, M.C., Filippi, J., Mencattini, A., Casti, P., Cerrato, G., Sauvat, A.,
  et~al.: Multi-scale generative adversarial network for improved evaluation of
  cell–cell interactions observed in organ-on-chip experiments. Neural.
  Comput. Appl.  (2020)

\bibitem{Goldsborough2017}
Goldsborough, P., Pawlowski, N., Caicedo, J.C., Singh, S., Carpenter, A.E.:
  Cytogan: generative modeling of cell images. BioRxiv p. 227645 (2017)

\bibitem{Goodfellow2014}
Goodfellow, I., Pouget-Abadie, J., Mirza, M., Xu, B., Warde-Farley, D., Ozair,
  S., Courville, A., Bengio, Y.: Generative adversarial nets. In: NeurIPS.
  vol.~27, pp. 2672--2680 (2014)

\bibitem{Hall2020}
Hall, M.S., Decker, J.T., Shea, L.D.: Towards systems tissue engineering:
  Elucidating the dynamics, spatial coordination, and individual cells driving
  emergent behaviors. Biomaterials  \textbf{255},  120189 (2020)

\bibitem{Han2018}
{Han}, L., {Murphy}, R.F., {Ramanan}, D.: Learning generative models of tissue
  organization with supervised gans. In: WACV. pp. 682--690 (2018)

\bibitem{Han2017}
{Han}, L., {Yin}, Z.: Transferring microscopy image modalities with conditional
  generative adversarial networks. In: CVPRW. pp. 851--859 (2017)

\bibitem{Henningsen2020}
Henningsen, J., Schwarz-Schilling, M., Leibl, A., Gutiérrez, J., Sagredo, S.,
  Simmel, F.C.: Single cell characterization of a synthetic bacterial clock
  with a hybrid feedback loop containing dcas9-sgrna. ACS Synth. Biol.
  \textbf{9}(12),  3377--3387 (2020)

\bibitem{Heusel2017}
Heusel, M., Ramsauer, H., Unterthiner, T., Nessler, B., Hochreiter, S.: {GANs}
  trained by a two time-scale update rule converge to a local nash equilibrium.
  In: NeurIPS. vol.~30, pp. 6626--6637 (2017)

\bibitem{Hofmann2019}
Hofmann, A., Falk, J., Prangemeier, T., Happel, D., K{\"{o}}ber, A.,
  Christmann, A., Koeppl, H., Kolmar, H.: {A tightly regulated and adjustable
  CRISPR-dCas9 based AND gate in yeast}. Nucleic Acids Res.  \textbf{47}(1),
  509--520 (2019)

\bibitem{Johnson2017}
Johnson, G.R., Donovan-Maiye, R.M., Maleckar, M.M.: Generative modeling with
  conditional autoencoders: Building an integrated cell. arXiv:1705.00092
  (2017)

\bibitem{Karras2020b}
Karras, T., Aittala, M., Hellsten, J., Laine, S., Lehtinen, J., Aila, T.:
  Training generative adversarial networks with limited data. In: NeurIPS.
  vol.~33, pp. 12104--12114 (2020)

\bibitem{Karras2019}
Karras, T., Laine, S., Aila, T.: A style-based generator architecture for
  generative adversarial networks. In: CVPR. pp. 4401--4410 (2019)

\bibitem{Karras2020a}
Karras, T., Laine, S., Aittala, M., Hellsten, J., Lehtinen, J., Aila, T.:
  Analyzing and improving the image quality of stylegan. In: CVPR. pp.
  8110--8119 (2020)

\bibitem{Kingma2015}
Kingma, D.P., Ba, J.: {Adam}: {A} method for stochastic optimization. In: ICLR
  (2015)

\bibitem{Lee2021}
Lee, G., Oh, J.W., Her, N.G., Jeong, W.K.: {DeepHCS++}: Bright-field to
  fluorescence microscopy image conversion using multi-task learning with
  adversarial losses for label-free high-content screening. Medical Image
  Analysis  \textbf{70},  101995 (2021)

\bibitem{Lee2018}
Lee, G., Oh, J.W., Kang, M.S., Her, N.G., Kim, M.H., Jeong, W.K.: Deephcs:
  Bright-field to fluorescence microscopy image conversion using deep learning
  for label-free high-content screening. In: MICCAI. pp. 335--343. Springer
  (2018)

\bibitem{Leygeber2019}
Leygeber, M., Lindemann, D., Sachs, C.C., Kaganovitch, E., Wiechert, W.,
  N{\"{o}}h, K., Kohlheyer, D.: {Analyzing Microbial Population Heterogeneity -
  Expanding the Toolbox of Microfluidic Single-Cell Cultivations}. J. Mol.
  Biol.  (2019)

\bibitem{Lugagne2020}
Lugagne, J., Lin, H., Dunlop, M.: {DeLTA: Automated cell segmentation,
  tracking, and lineage reconstruction using deep learning}. PLOS Comput. Biol.
   \textbf{16}(4) (2020)

\bibitem{Mescheder2018}
Mescheder, L., Geiger, A., Nowozin, S.: Which training methods for {GANs} do
  actually converge? In: ICML. pp. 3481--3490 (2018)

\bibitem{Osokin2017}
Osokin, A., Chessel, A., Carazo~Salas, R.E., Vaggi, F.: Gans for biological
  image synthesis. In: ICCV (2017)

\bibitem{Paszke2019}
Paszke, A., Gross, S., Massa, F., Lerer, A., Bradbury, J., et~al.: {PyTorch}:
  An imperative style, high-performance deep learning library. In: NeurIPS.
  vol.~32, pp. 8026--8037 (2019)

\bibitem{Pepperkok2006}
Pepperkok, R., Ellenberg, J.: High-throughput fluorescence microscopy for
  systems biology. Nat. Rev. Mol. Cell Biol. p.~690 (2006)

\bibitem{Prangemeier2020b}
{Prangemeier}, T., {Wildner}, C., {Françani}, A.O., {Reich}, C., {Koeppl}, H.:
  Multiclass yeast segmentation in microstructured environments with deep
  learning. In: IEEE CIBCB. pp.~1--8 (2020)

\bibitem{Prangemeier2020}
Prangemeier, T., Lehr, F.X., Schoeman, R.M., Koeppl, H.: {Microfluidic
  platforms for the dynamic characterisation of synthetic circuitry}. Curr.
  Opin. Biotechnol.  \textbf{63},  167--176 (2020)

\bibitem{Prangemeier2020c}
Prangemeier, T., Reich, C., Koeppl, H.: Attention-based transformers for
  instance segmentation of cells in microstructures. In: IEEE BIBM. pp.
  700--707 (2020)

\bibitem{Prangemeier2018}
Prangemeier, T., Wildner, C., Hanst, M., Koeppl, H.: {Maximizing information
  gain for the characterization of biomolecular circuits}. In: Proc. 5th
  ACM/IEEE NanoCom. pp.~1--6 (2018)

\bibitem{Riba2020}
Riba, E., Mishkin, D., Ponsa, D., Rublee, E., Bradski, G.: Kornia: an open
  source differentiable computer vision library for {PyTorch}. In: WACV. pp.
  3663--3672 (2020)

\bibitem{Salimans2016}
Salimans, T., Goodfellow, I., Zaremba, W., Cheung, V., Radford, A., Chen, X.,
  Chen, X.: Improved techniques for training {GANs}. In: NeurIPS. vol.~29, pp.
  2234--2242 (2016)

\bibitem{Schonfeld2020}
Schonfeld, E., Schiele, B., Khoreva, A.: A {U-Net} based discriminator for
  generative adversarial networks. In: CVPR. pp. 8207--8216 (2020)

\bibitem{Sinha2020}
Sinha, S., Zhao, Z., Goyal, A., Raffel, C.A., Odena, A.: Top-k training of
  {GANs}: Improving gan performance by throwing away bad samples. In: NeurIPS.
  vol.~33, pp. 14638--14649 (2020)

\bibitem{Szegedy2016}
Szegedy, C., Vanhoucke, V., Ioffe, S., Shlens, J., Wojna, Z.: Rethinking the
  inception architecture for computer vision. In: CVPR. pp. 2818--2826 (2016)

\bibitem{Ulman2016}
Ulman, V., Svoboda, D., Nykter, M., Kozubek, M., Ruusuvuori, P.: Virtual cell
  imaging: A review on simulation methods employed in image cytometry.
  Cytometry Part A  \textbf{89}(12),  1057--1072 (2016)

\bibitem{Unterthiner2019}
Unterthiner, T., van Steenkiste, S., Kurach, K., Marinier, R., Michalski, M.,
  Gelly, S.: {FVD:} {A} new metric for video generation. In: ICLR Workshop
  (2019)

\bibitem{Wang2020}
Wang, N.B., Beitz, A.M., Galloway, K.: Engineering cell fate: Applying
  synthetic biology to cellular reprogramming. Curr. Opin. Syst. Biol.
  \textbf{24},  18--31 (2020)

\bibitem{Wang2018}
Wang, X., Girshick, R., Gupta, A., He, K.: Non-local neural networks. In: CVPR.
  pp. 7794--7803 (2018)

\bibitem{Wieslander2021}
Wieslander, H., Gupta, A., Bergman, E., Hallstr{\"o}m, E., Harrison, P.J.:
  Learning to see colours: generating biologically relevant fluorescent labels
  from bright-field images. bioRxiv  (2021)

\bibitem{Zhang2019}
Zhang, H., Fang, C., Xie, X., Yang, Y., Mei, W., Jin, D., Fei, P.:
  High-throughput, high-resolution deep learning microscopy based on
  registration-free generative adversarial network. Biomedical optics express
  \textbf{10}(3),  1044--1063 (2019)

\end{thebibliography}
	
	

\section*{Supplementary material}

# Multi-StyleGAN: Towards Image-Based Simulation of Time-Lapse Live-Cell Microscopy – Supplementary Materials –

Christoph Reich*, Tim Prangemeier*, Christian Wildner, and Heinz Koeppl

`heinz.koeppl@bcs.tu-darmstadt.de`

**Table S1.** Multi-StyleGAN hyperparameters overview.

| Hyperparameter | Value |
|---|---|
| Training epochs | 100 |
| Generator/Discriminator lazy regularization | every 16 training step |
| $R_1$ weight | 10 |
| CutMix augmentation weight | 4 |
| Consistency regularization weight | 4 |
| Path length regularization weight | $\log 2 / (256^2 \cdot (\log 256 - \log 2))$ |
| Batch size path length regularization | 12 |
| Batch size | 24 |
| Generator learning rate | $2 \cdot 10^{-4}$ |
| Discriminator learning rate | $6 \cdot 10^{-4}$ |
| Mapping network learning rate | $2 \cdot 10^{-6}$ |
| Optimiser | Adam ($\beta_1 = 0$, $\beta_2 = 0.99$) |
| Top-$k$ samples (after annuling) | 12 |
| Top-$k$ linear annealing start epoch | 25 |
| Top-$k$ linear annealing finish epoch | 75 |
| Disordered sequences start epoch | 75 |
| Disordered sequences batch size | 6 |
| Input noise vector shape | $\mathbb{R}^{512}$ sampled from $\mathcal{N}(0, 1)$ |
| Mapping network depth | 8 |
| Mapping network features | 512 |
| Generator conv. features per stage | 512, 512, 512, 512, 512, 512, 512 |
| Discriminator encoder conv. features per stage | 128, 256, 384, 768, 1024 |
| Discriminator decoder conv. features per stage | 768, 384, 256, 128 |
| Discriminator scale prediction mapping | two-layer feed forward neural network |
| Adaptive discriminator augmentations | pixel blitting & geometric transforms |
| Adaptive discriminator $p$ step size | $5 \cdot 10^{-3}$ every 8 training step |
| Adaptive discriminator $r$ target | 0.6 |
| Generator parameters | 51.01M |
| Discriminator parameters | 50.85M |

---

* Christoph Reich and Tim Prangemeier — both authors contributed equally

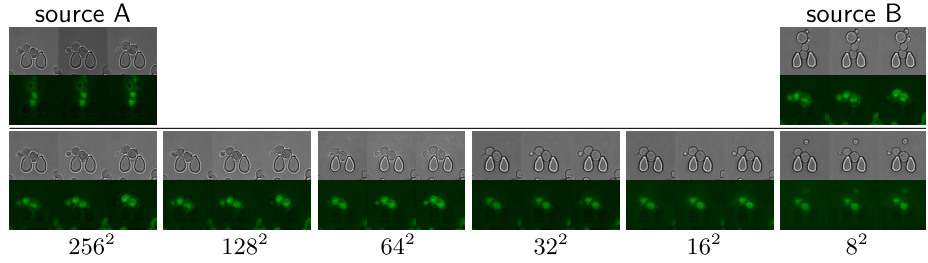

**Fig. S1.** Style-mixing example between source sequences A and B. The numbers correspond to the resolution of the stage into which the latent vector of source sample B is employed for the subsequent stages. BF in grayscale and GFP channels below in green. We recommend zooming in to inspect the sample quality in detail.

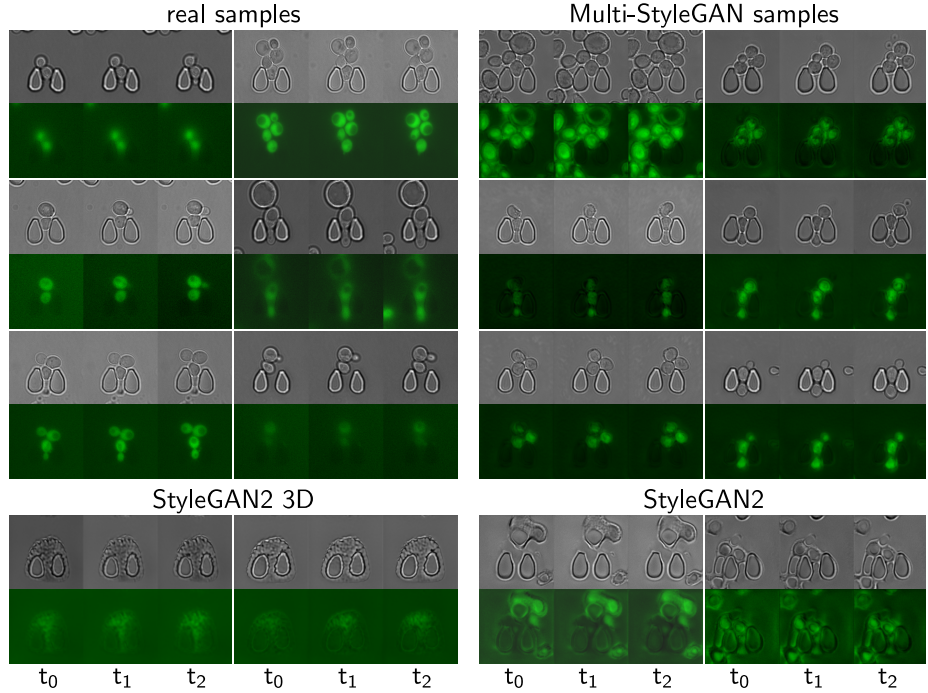

**Fig. S2.** Real BF & GFP sequences of the trapped yeast cell dataset over three timesteps (top three rows on left). Multi-StyleGAN BF & GFP sequences (top three rows on right). The original StyleGAN2 models the temporal dimension as well as BF and GFP in the channel dimension. Samples of StyleGAN2 with ADA and a U-Net discriminator (bottom right). StyleGAN2 3D uses 3D convolutions, modeling the temporal dimension in the convolution itself. Samples of StyleGAN2 3D with ADA and a U-Net discriminator (bottom left). The hyparameters listed in Table S1 were employed for both StyleGAN2 and StyleGAN2 3D.


	
\end{document}